\documentclass[]{academic_template}

\usepackage[toc,page,header]{appendix}

\usepackage{marvosym}

\usepackage{amssymb}
\usepackage{amsmath}

\usepackage{makecell}
\usepackage{algorithm}      
\usepackage{algpseudocode}  

\usepackage{graphicx}

\usepackage{listings}
\usepackage{url}
\usepackage{booktabs}
\usepackage{wrapfig}

\usepackage{colortbl}
\usepackage{xcolor}

\usepackage{caption}
\usepackage{subcaption}

\newcommand{\Rmnum}[1]{\expandafter\@slowromancap\romannumeral #1@}

\usepackage{bbding}

\usepackage{amssymb}
\usepackage{pifont}
\usepackage{multirow}
\usepackage[table,xcdraw]{xcolor}
\useunder{\uline}{\ul}{}

\setlogoheight{14mm}   
\setlogospacing{5mm}
\setsjtublue 

\settitlerulethickness{3pt}  

\abstractboxon 
\setabstractframecolor{gray} 
\setabstractbgcolor{gray!10} 
\setlogotolineshift{5mm} 

\settoprulethickness{2.5pt}  
\setbottomrulethickness{1.5pt} 

\newcommand{\modelname}{EvoTok\xspace}

\usepackage{xspace}
\usepackage{multirow}
\usepackage[table,xcdraw]{xcolor}
\usepackage{multirow}

\usepackage{graphicx}
\usepackage{subcaption}

\title{\modelname: A Unified Image Tokenizer via Residual Latent Evolution for Visual Understanding and Generation}

\author[1, *]{Yan Li}
\author[1, *\dagger]{Ning Liao}
\author[1]{Xiangyu Zhao}
\author[2]{Shaofeng Zhang}
\author[1]{Xiaoxing Wang}
\author[3]{Yifan Yang}
\author[1]{Junchi Yan}
\author[1, \text{\Letter}]{Xue Yang}

\affiliation[1]{Shanghai Jiao Tong Univerisity}
\affiliation[2]{University of Science and Technology of China}
\affiliation[3]{Microsoft Corporation}

\contribution[*]{Equal Contributors}
\contribution[\dagger]{Project Leader}
\contribution[\text{\Letter}]{Corresponding Author}

\abstract{
The development of unified multimodal large language models (MLLMs) is fundamentally challenged by the granularity gap between visual understanding and generation: understanding requires high-level semantic abstractions, while image generation demands fine-grained pixel-level representations. Existing approaches usually enforce the two supervision on the same set of representation or decouple these two supervision on separate feature spaces, leading to interference and inconsistency, respectively. In this work, we propose \textbf{\modelname}, a unified image tokenizer that reconciles these requirements through a residual evolution process within a shared latent space. Instead of maintaining separate token spaces for pixels and semantics, \modelname encodes an image into a cascaded sequence of residual tokens via residual vector quantization. 
This residual sequence forms an evolution trajectory where earlier stages capture low-level details and deeper stages progressively transition toward high-level semantic representations.
Despite being trained on a relatively modest dataset of 13M images, far smaller than the billion-scale datasets used by many previous unified tokenizers, \modelname achieves a strong reconstruction quality of 0.43 rFID on ImageNet-1K at 256$\times$256 resolution. When integrated with a large language model, \modelname shows promising  performance across 7 out of 9 visual understanding benchmarks, and remarkable results on image generation benchmarks such as GenEval and GenAI-Bench. These results demonstrate that modeling visual representations as an evolving trajectory provides an effective and principled solution for unifying visual understanding and generation.
}

\date{\today}

\checkdata[Code]{\url{https://github.com/VisionXLab/EvoTok}}

\begin{document}
\maketitle


\section{Introduction}

\begin{figure}[t!]
    \centering
    \includegraphics[width=\linewidth]{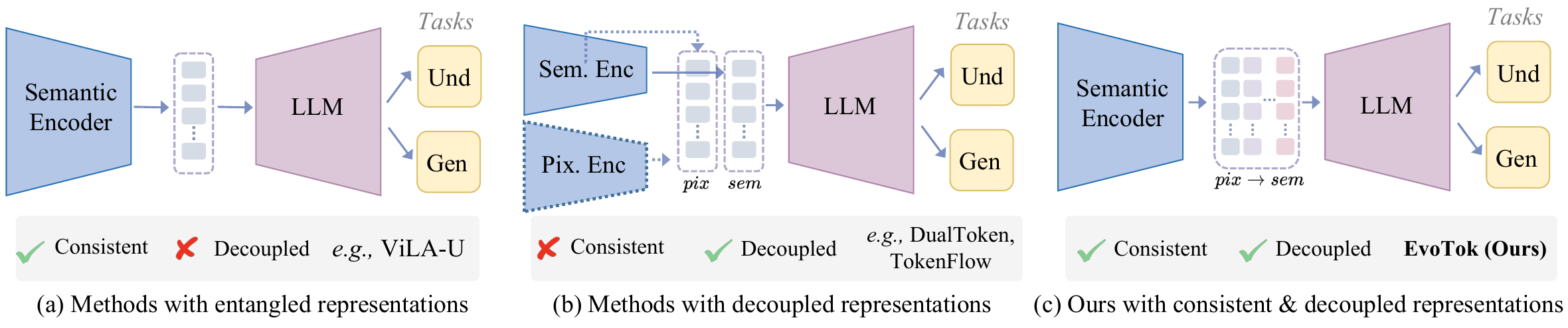}
    \caption{\textbf{Different paradigms of current MLLMs and unified models.}
(a) Entangled unified MLLMs share semantic and pixel features, enabling both understanding and generation but tightly coupling the two objectives.
(b) Decoupled unified MLLMs separate semantic and pixel encoders or feature layers to disentangle tasks.
(c) Our unified evolution MLLM evolves pixel features into semantic representations within a shared space while keeping decoupled latents, achieving aligned understanding and generation while preserving decoupled representations.
}
    \label{fig:teaser}
\end{figure}

Recent advances in Multimodal Large Language Models (MLLMs)~\cite{liu2023llava,lin2024vila,bai2023qwenvl,chen2025blip3} have revolutionized visual reasoning, while diffusion models~\cite{podell2023sdxl,rombach2022sdv15_21} and autoregressive frameworks~\cite{sun2024llamagen,tian2024var} have enabled high-fidelity image synthesis. Although these paradigms have evolved separately, their complementary nature has sparked growing interest for unified visual understanding and generation, the core of which is designing a unified image tokenizer supporting both tasks. 

Specifically, one line of work~\cite{wu2024vilau,ma2025unitok} unifies the text-image contrastive loss and VQ-based image reconstruction loss upon the same set of visual representations extracted by the designed tokenizers for consistency, as shown in Fig.~\ref{fig:teaser} (a). This design forces the representation to serve as a bridge, capturing both the high-level semantics essential for understanding and the low-level pixel fidelity required for generation. However, it introduces an inherent optimization conflict: understanding tasks prioritize semantic alignment and discriminability, while generation tasks focus on fine-grained reconstruction and structural fidelity. These competing objectives interfere with each other within the shared feature space, undermining the representation's effectiveness for either task. Therefore, explicitly decoupling the features for these two tasks is essential.

To achieve feature decoupling, another line of approaches~\cite{song2025_dualtoken,lin2025toklip,chen2025semhitok,geyer2023tokenflow} learns semantic-wise features for understanding and pixel-wise features for generation independently, as shown in Fig.~\ref{fig:teaser} (b). Nonetheless, this overly independent decoupling directly compromises the intrinsic consistency and correlation between the two types of features. The understanding and generation branches become isolated in the feature space, failing to share visual structures and semantic priors, which is detrimental to effective modeling within a unified architecture.

Building on these analysis, we advocate for a holistic design principle: \textit{a unified tokenizer should not only decouple representations to mitigate task conflicts, but also preserve consistency between them to facilitate the sharing of visual appearance and semantic information}. In this work, we address both objectives by evolving pixel-level features into semantic-level features within a shared space.

We propose \textbf{\modelname}, a unified image tokenizer that achieves task-decoupled and consistent visual representations through a \textit{residual evolution process}, as shown in Fig.~\ref{fig:teaser} (c). It leverages cascaded residual vector quantization to decompose an image into a sequence of residual tokens within a \textit{shared latent space}. Specifically, earlier residual stages capture spatial structures and perceptual details that are critical for image reconstruction, while deeper stages gradually steer the representation toward semantic features aligned with pretrained vision-language models. By organizing the visual representation along this residual trajectory, our tokenizer naturally reconciles the representational requirements of both tasks: pixel-level features are preserved for high-fidelity generation, while the accumulated representation encodes semantic abstractions suitable for visual understanding. Importantly, both forms of information naturally coexist within a single, unified latent space for task-consistency, rather than being maintained in separate ones.

Experimentally, despite being trained on a significantly smaller data scale with 13M images compared to baselines using billion-scale training data, the proposed \modelname achieves a strong rFID score of 0.43 at 256$\times$256 resolution on ImageNet-1K~\cite{deng2009imagenet}. Moreover, \modelname contributes to the best performance on 7 out of 9 visual understanding tasks (e.g., SEEDBench~\cite{seedbench}, MMMU~\cite{yue2024mmmu}, and MME~\cite{fu2023mme}), and dominating complex image generation benchmarks, including GenEval~\cite{geneval} and GenAI-Bench~\cite{jiang2024genai}.

Our contributions are summarized as follows:

(1) We propose \modelname, a unified image tokenizer that represents images as a residual latent evolution trajectory within a shared latent space.

(2) We introduce a unified training objective that strikes a balance between task decoupling and cross-task consistency, learning visual representations that are simultaneously effective for semantic alignment and pixel reconstruction.

(3) Extensive experiments demonstrate the superior performance achieved by the proposed \modelname on both visual understanding and generation benchmarks, and its effectiveness has been comprehensively validated.

\section{Related Work}

\subsection{Image Tokenizer for Understanding}

Image tokenization is a fundamental component of multimodal large language models (MLLMs)~\cite{liu2023llava,lin2024vila,bai2023qwenvl,chen2025blip3}, which require converting images into compact visual tokens that can be processed together with text tokens.
A common design is to adopt pretrained vision encoders that produce semantically rich visual representations.
For instance, CLIP~\cite{radford2021clip} learns image representations aligned with text through large-scale contrastive pretraining, making it a widely adopted visual tokenizer for MLLMs~\cite{liu2023llava}.
Self-supervised visual encoders such as DINOv2~\cite{oquab2023dinov2} have also been explored due to their strong performance on region-level and dense visual understanding tasks~\cite{ma2024groma}.
While these continuous tokenizers excel at capturing abstract concepts, their continuous latent spaces present significant hurdles for unified modeling alongside discrete text tokens in autoregressive frameworks. To bridge this gap, recent studies have attempted to discretize continuous semantic features~\cite{ge2023seed} or directly employ discrete, reconstruction-based tokenizers~\cite{liu2024world} like VQVAE~\cite{van2017vqvae}.

\subsection{Image Tokenizer for Generation}

In visual generation, image tokenizers aim to encode images into compact latent tokens that preserve detailed spatial information for high-quality reconstruction and generation.
The seminal VQVAE~\cite{van2017vqvae} introduced discrete tokenization by mapping continuous latent features to a learnable codebook, enabling images to be represented as sequences of discrete tokens. 
Such vector-quantized tokenizers are widely adopted due to their discrete latent space and compatibility with autoregressive and masked generative models~\cite{sun2024llamagen,chang2022maskgit,tian2024var}. 
Subsequent improvements further enhanced reconstruction fidelity. 
VQGAN~\cite{yu2021vqgan} introduced adversarial and perceptual losses to improve synthesis quality, while later works explored more expressive quantization strategies such as residual quantization~\cite{lee2022_rqvae}. Despite their strong reconstruction capability, these tokenizers are primarily optimized for pixel-level reconstruction and tend to focus on low-level visual details, often lacking strong semantic alignment with language representations. Consequently, models built upon such tokenizers often struggle with visual understanding tasks.

\subsection{Unified Image Tokenizer}
The pursuit of native multimodal systems has driven recent efforts to unify visual understanding and generation within a single architecture~\cite{wu2024vilau,wang2024emu3,wu2025janus,lu2024unifiedio2}. A key challenge in this direction lies in designing an image tokenizer that can simultaneously support semantic reasoning and pixel-level generation. Existing unified tokenization strategies generally fall into two paradigms. One line of work adopts an \textbf{\textit{entangled}} representation design, where both understanding and generation objectives are optimized on the same visual features. For example, models such as ViLA-U~\cite{wu2024vilau} and UniTok~\cite{ma2025unitok} integrate image reconstruction losses~\cite{yu2021vqgan} with language alignment objectives~\cite{radford2021clip} based on quantized features. Although such designs allow both tasks to share a unified representation space, they suffer from an inherent optimization conflict between semantic alignment and fine-grained reconstruction, ultimately limiting the upper-bound performance of both tasks.

To solve this, another line of work attempts to explicitly \textit{\textbf{decouple}} semantic and pixel representations. 
Several strategies have been explored to achieve this goal. 
Some methods separate these two representations across different feature layers within the tokenizer, allowing early layers to capture pixel-level details while deeper layers focus on semantic abstractions~\cite{song2025_dualtoken,lin2025toklip}. 
Other approaches introduce dedicated branches to independently model semantic and reconstruction features~\cite{chen2025semhitok,geyer2023tokenflow}. 
Additionally, some works employ separate codebooks to store semantic and pixel tokens, explicitly partitioning the latent space for different objectives~\cite{chen2025semhitok,song2025_dualtoken}. 
While these designs alleviate optimization interference between tasks, the resulting representations often become overly independent, weakening the intrinsic consistency between visual structure and semantic information.

\section{Method} 
 
In this section, we introduce \modelname, a unified image tokenizer for visual understanding and generation that learns both semantic-level and pixel-level representations through residual evolution within a shared latent space. We first present the overall framework (Sec.~\ref{sec:method_tok}), followed by its integration into visual understanding (Sec.~\ref{sec:method_und}) and generation (Sec.~\ref{sec:method_gen}) tasks.

\subsection{Unified Image Tokenizer}
\label{sec:method_tok}

\begin{figure}[t!]
    \centering
    \includegraphics[width=0.95\linewidth]{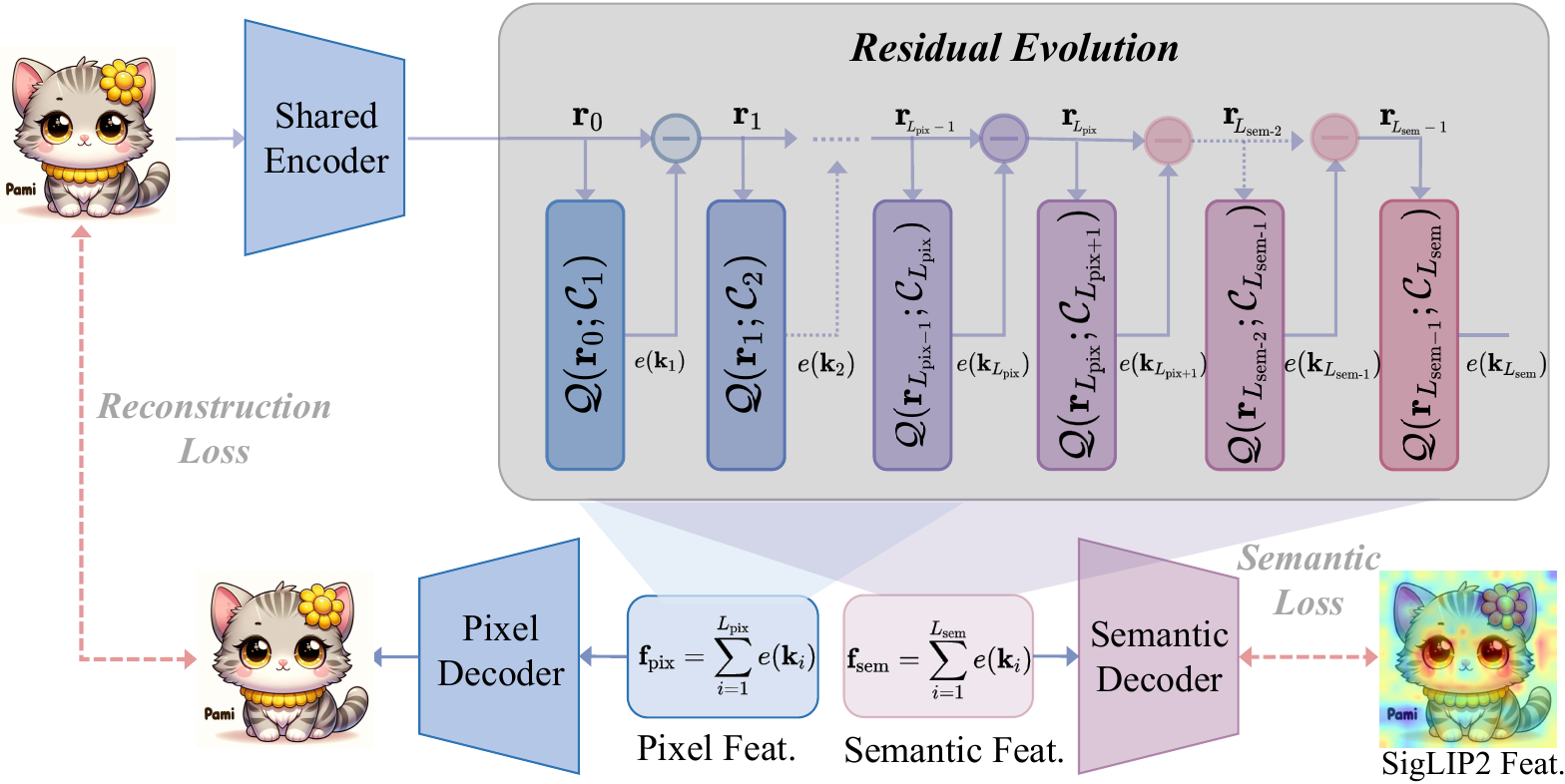}
    \caption{\textbf{The overview of the proposed \modelname.} An image is encoded into a sequence of residual tokens within a shared latent space, forming a residual evolution trajectory. Tokens from earlier stages preserve \textbf{pixel-level} details for reconstruction, while tokens from deeper stages, formed by progressively accumulating tokens from full stages, capture \textbf{semantic-level} features aligned with those extracted by the pre-trained SigLIP2~\cite{tschannen2025siglip2}. Within this residual evolution trajectory, low-level pixel features and high-level semantic representations \textit{\textbf{co-evolve}}, enabling a decoupled and consistent representations to support both visual understanding and generation.}
    \label{fig:framework}
\end{figure}

We propose \textbf{\modelname}, which represents an image as a residual latent evolution trajectory within a shared feature space. This design simultaneously achieves task-decoupled features for understanding and generation, while maintaining consistency between pixel-level and semantic-level representations. Along with the evolution process, earlier residual stages capture structural and perceptual features for low-level reconstruction, while deeper stages progressively accumulate and refine prior features into high-level semantic tokens. By sharing this trajectory across both tasks, \modelname achieves a truly unified representation where pixels and semantics co-evolve, naturally reconciling the demands of generation and understanding within a single latent space. The overview of the proposed \textbf{\modelname} is presented in Fig.~\ref{fig:framework}.

\noindent
\textbf{Residual Evolution within Shared Latent Space.} 
We propose the residual evolution strategy to tokenize an image $I$ into task-decoupled and cross-task consistent representations, maintaining pixel-level and semantic-level features for visual generation and understanding. Given an initial feature $\mathbf{f}$ extracted by the shared encoder $\mathcal{E}$, a cascaded residual vector quantization composed of $L$ stages is performed following the RQ-VAE~\cite{lee2022_rqvae}, which is formulated as:
\begin{equation}
(\mathbf{k}_1, \dots, \mathbf{k}_L) = \mathcal{RQ}(\mathbf{f}; \{\mathcal{C}_i\}_{i=1}^L),
\end{equation}
where $\mathcal{C}_i$ is the codebook for stage $i$. Setting the initial input $\mathbf{r}_0$ of the residual quantization process as the continuous feature $\mathbf{r}_0=\mathbf{f}$, each stage recursively quantizes the residual as:
\begin{align}
\mathbf{k}_i = \mathcal{Q}(\mathbf{r}_{i-1}; \mathcal{C}_i), \quad \mathbf{r}_i = \mathbf{r}_{i-1} - \mathbf{e}_i(\mathbf{k}_i), \quad i = 1, \dots, L.
\end{align}

Here, $\mathbf{e}_i$ is the embedding table of $\mathcal{C}_i$, and $\mathcal{Q}$ is the standard vector quantization process formulated as:
\begin{equation}
\mathcal{Q}(\mathbf{z}; \mathcal{C}) = \arg \min_{k \in [K]} \|\mathbf{z} - \mathbf{e}(k)\|_2^2,
\end{equation}
in which $K=|\mathcal{C}|$ denotes the codebook size.

Along the residual quantization process, we propose the \textit{pixel-to-semantic} progression: early residuals capture perceptual structure for high-fidelity reconstruction, while the accumulated trajectory converges to high-level semantics. We define two pivotal features along the residual trajectory. The pixel-level feature $\mathbf{f}_\text{pix}$ is computed as the partial sum of earlier stages, while the semantic feature $\mathbf{f}_\text{sem}$ is the cumulative sum over the full trajectory:
\begin{equation}
\label{equ:features}
\mathbf{f}_\text{pix} = \sum_{i=1}^{L_\text{pix}} \mathbf{e}_i(\mathbf{k}_i), \quad
\mathbf{f}_\text{sem} = \sum_{i=1}^{L_\text{sem}} \mathbf{e}_i(\mathbf{k}_i),
\end{equation}
where $L_\text{pix}$ and $L_\text{sem}$ denote the number of earlier residual stages used for the pixel-level representation and the total number of residual stages in the trajectory, respectively. The pixel feature $\mathbf{f}_\text{pix}$ preserves dominant structures and dense perceptual details for generation, and the semantic feature $\mathbf{f}_\text{sem}$ represents refined semantics aligned with SigLIP2~\cite{tschannen2025siglip2}. 

\noindent
\textbf{Unified Training Objective.}
To enable the features extracted along the evolution trajectory to simultaneously support visual generation and understanding, we propose a unified training objective that enforces both pixel-level reconstruction and semantic alignment. In the \textit{pixel-to-semantic} residual evolution process, the pixel-level feature $\mathbf{f}_\text{pix}$ is decoded by a pixel decoder for reconstruction, and the semantic feature $\mathbf{f}_\text{sem}$ is decoded by a semantic decoder to align with the feature extracted by SigLIP2 model. The total training objective is: 
\begin{equation}
\mathcal{L}_\text{total} = \mathcal{L}_\text{pix} + \mathcal{L}_\text{sem} + \mathcal{L}_\text{VQ},
\end{equation}
where the semantic loss $\mathcal{L}_\text{sem}$ is defined as the cosine similarity between the decoded semantic feature and target SigLIP2 embeddings. The pixel-level loss consists of reconstruction loss $\mathcal{L}_\text{R}$, perceptual loss $\mathcal{L}_\text{P}$, and adversarial loss $\mathcal{L}_\text{G}$:
\begin{equation}
\mathcal{L}_\text{pix} = \mathcal{L}_\text{R} + \lambda_\text{P} \mathcal{L}_\text{P} + \lambda_\text{G}\mathcal{L}_\text{G}.
\end{equation}

In addition, the residual codebooks are trained with a standard VQ loss~\cite{yu2021vqgan}:
\begin{equation}
\mathcal{L}_\text{VQ} = \sum_{d=1}^{L} \|\mathbf{f}-\text{sg}[\widehat{\mathbf{f}}_d]\|_2^2,
\end{equation}
where $\widehat{\mathbf{f}}_d = \sum_{i=1}^{d} \mathbf{e}_i(\mathbf{k}_i)$ is the $d$-th layer feature from the residual trajectory, $\text{sg}[\cdot]$ denotes stop-gradient.

\subsection{Visual Understanding with \modelname}
\label{sec:method_und}

With the unified visual tokenizer aligned with language representations, we build a MLLM for visual understanding following the LLaVA~\cite{liu2023llava} paradigm. 
Specifically, \modelname\ encodes an image into $H \times W \times L_\text{sem}$ discrete codes, where $L_\text{sem}$ denotes the depth of the residual trajectory.
For each spatial location, the semantic feature $\mathbf{f}_\text{sem}$ is obtained by aggregating residual embeddings along the trajectory as in Eq.(\ref{equ:features}). These semantic features are decoded by the semantic decoder and projected into the language embedding space through an MLP projector, which are then fed into the LLM together with text tokens. The model is trained with the standard autoregressive language modeling objective:
\begin{equation}
\mathcal{L}_\text{CE} = - \sum_{t} \log p(y_t \mid y_{<t}, \mathbf{v}),
\end{equation}
where $\mathbf{v}$ denotes the visual tokens derived from \modelname.

\subsection{Visual Generation with \modelname}
\label{sec:method_gen}

For image generation, the LLM autoregressively predicts the pixel residual trajectory codes produced by the unified tokenizer.  Following ViLA-U~\cite{wu2024vilau} and UniTok~\cite{ma2025unitok}, we employ a depth RQ-Transformer~\cite{lee2022_rqvae} head to generate $L_\text{pix}$ codes for each spatial token.
Specifically, given the hidden representation $\mathbf{h}_{s}$ from the LLM at spatial position $s$, the RQ head autoregressively predicts the residual codes $(\hat{\mathbf{k}}_{s,1}, \dots, \hat{\mathbf{k}}_{s,L_\text{pix}})$ along the depth dimension. At each depth level $i \in \{1, \dots, L_\text{pix}\}$, the code is sampled from a categorical distribution:
\begin{equation}
\hat{\mathbf{k}}_{s,i} \sim P(\mathbf{k}_{s,i} \mid \mathbf{h}_{s}, \hat{\mathbf{k}}_{s,<i}),
\end{equation}
where a dedicated classifier is used for each codebook level $\mathcal{C}_i$.

To reconstruct the image, the predicted codes are mapped back to their feature embeddings $\{\mathbf{e}_i(\hat{\mathbf{k}}_{s,i})\}_{i=1}^{L_\text{pix}}$. These embeddings are accumulated along the residual trajectory to retrieve the pixel-level feature $\hat{\mathbf{f}}_{\text{pix}, s}$ as defined in Eq.~(\ref{equ:features}):
\begin{equation}
\hat{\mathbf{f}}_{\text{pix}, s} = \sum_{i=1}^{L_\text{pix}} \mathbf{e}_i(\hat{\mathbf{k}}_{s,i}).
\end{equation}
The resulting feature map $\hat{\mathbf{f}}_\text{pix}$ is then processed by the pixel decoder to generate the final image. During training, the generation model is optimized using a cross-entropy loss over the ground-truth residual codes:
\begin{equation}
\mathcal{L}_\text{gen} = - \sum_{s} \sum_{i=1}^{L_\text{pix}} \log P(\mathbf{k}_{s,i} \mid \mathbf{h}_{s}, \mathbf{k}_{s,<i}).
\end{equation}

\section{Experiments}
In this section, we first introduce the experimental setup (Sec.~\ref{sec:exp_setup}), including the datasets, implementation details, and evaluation metrics.
We then present quantitative and qualitative results on reconstruction (Sec.~\ref{sec:result_rec}), understanding, and generation benchmarks (Sec.~\ref{sec:result_mllm}).
Finally, we conduct ablation studies to analyze the effectiveness of the proposed pixel-to-semantic evolution (Sec.~\ref{sec:resule_ablation}).

\subsection{Experimental Setup}
\label{sec:exp_setup}
\noindent
\textbf{Datasets.} We utilize \textbf{\textit{fully open-sourced data without any re-caption or distillation}} across all of the training stages of our model. Specifically, 
\modelname is trained on CC12M~\cite{changpinyo2021cc12m} and ImageNet-1K~\cite{deng2009imagenet} trainset, and evaluated on ImageNet-1K valset. For multimodal understanding, LLaVA-Pretrain-558\textit{k}~\cite{liu2023llava} is used for pretraining, and 13M data sampled from LLaVA-OneVision~\cite{lillavaov}, LLaVA-NeXT-780\textit{k}~\cite{liu2024llavanext}, Cambrian-10M~\cite{tong2024cambrian}, and HoneyBee~\cite{bansal2025honeybee} are used for supervised finetuning (SFT). For image generation,  we sample 22M data from  ImageNet1K-QwenImage~\cite{han2025imagenet1k_qwenimage}, BLIP3o-Pretrain-Short~\cite{chen2025blip3}, LLaVA-OneVision-1.5~\cite{LLaVA-OneVision-1.5}, ShareGPT-4o-Image~\cite{chen2025sharegpt4oimg} and OpenGPT-4o-Image~\cite{chen2025opengpt4o}. While previous unified tokenizers~\cite{geyer2023tokenflow,ma2025unitok,lin2024vila} are often trained on billion-scale image–text pairs such as COYO-700M~\cite{kakaobrain2022coyo-700m} or DataComp-1B~\cite{gadre2023datacomp}, our model is trained on a significantly smaller scale and still achieves even stronger performance.

\noindent
\textbf{Implementation Details.}
We employ SigLip2-large-patch16-256~\cite{tschannen2025siglip2} as our vision encoder architecture and the teacher model. Residual quantization~\cite{lee2022_rqvae} is adopted to produce the codes, with each codebook containing 32,768 entries for each depth. All images are resized to $256\times256$ and encoded into $16\times16\times L$ tokens, where the first $L_\text{pix}=4$ depths of tokens preserve pixel-level features, while the full $L_\text{sem}=L=16$ depths accumulated into semantic-level representations. The tokenizer is trained for 500$k$ steps with a global batch size of 256, and the learning rate is set to 1$e^{-4}$.
For MLLM, we choose Qwen-2.5-7B-Instruct~\cite{qwen2.5} as the language model and a depth transformer as in~\cite{wu2024vilau} as the generation head. All training experiments are performed on 16 GPUs. At inference, classifer-free guidance~\cite{ho2022classifier} is applied for image generation with a scale factor of 7.5.

\noindent
\textbf{Evaluation Metrics.}
We evaluate reconstruction quality using Fr\'echet Image Distance (FID)~\cite{heusel2017gans} on ImageNet-1K validation set~\cite{deng2009imagenet}. For multimodal understanding, we conduct evaluations across a comprehensive suite of vision-language benchmarks, including SEEDBench \cite{seedbench}, MM-Vet \cite{mmvet}, GQA \cite{gqa}, AI2D \cite{ai2d}, RealWorldQA \cite{realworldqa}, MMMU \cite{yue2024mmmu}, MMBench \cite{liu2025mmbench}, MME \cite{fu2023mme} and MME-P (MME-Perception). Besides, the image generation capability is evaluated on GenAI-Bench~\cite{jiang2024genai} and GenEval~\cite{geneval}. Following~\cite{ramesh2022dalle3,wang2024emu3,geyer2023tokenflow}, we report our GenEval results using GPT-4o as a rewriter.
\begin{figure}[t!]
    \centering
    \includegraphics[width=\linewidth]{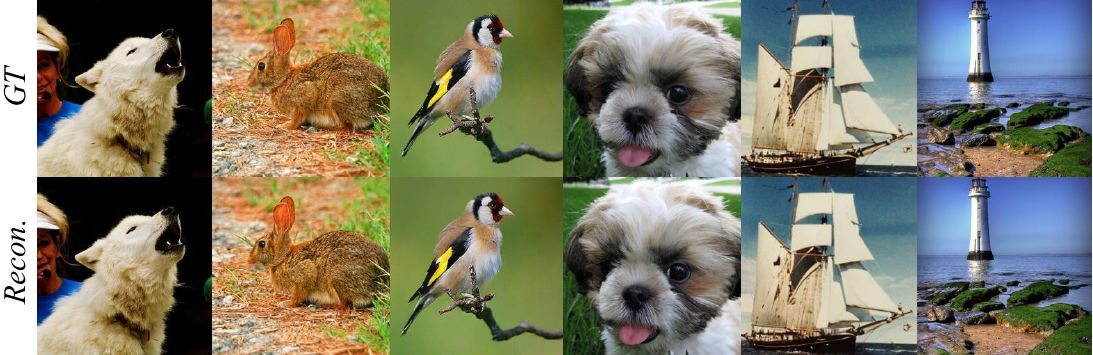}
    \caption{\textbf{Images reconstruction at a resolution of 256$\times$ 256 on ImageNet-1K.} }
    \label{fig:reconstruction_figs}
\end{figure}
\begin{table}[t!]
\centering
\caption{\textbf{Comparison of the reconstruction FID on ImageNet-1K.} rFID is measured at 256$\times$256 resolution. \textbf{Bold} and \underline{underline} indicate the best and the second best performance, respectively. $\dagger$: indicates model trained on \textit{\textbf{billion-scale}} datasets such as COYO-700M~\cite{kakaobrain2022coyo-700m} or DataComp-1B~\cite{gadre2023datacomp}.}
\label{tab:main_rfid}
\begin{tabular}{lccccc}
\toprule
\textbf{Model} &  \textbf{Resolution} & \quad\textbf{Code Shape}\quad & \quad\textbf{Codebook Size}\quad & \quad\textbf{rFID$\downarrow$}\quad \\ \hline \rowcolor[HTML]{EFEFEF} 
\multicolumn{5}{l}{\textit{\textbf{Reconstruction Only}}} \\
LLaMAGen~\cite{sun2024llamagen} & 256 & 16$\times$16 & 16,384 & 2.19 \\
RQVAE~\cite{lee2022_rqvae} & 256 & 16$\times$16$\times$4 & 16,384 & 3.20 \\
VQGAN-LC~\cite{zhu2024_vqganlc} & 256 & 16$\times$16 & 16,384 & 2.62 \\
IBQ~\cite{shi2025_ibq} & 256 & 16$\times$16 & 252,144 & 1.00 \\ \hline\rowcolor[HTML]{EFEFEF} 
\multicolumn{5}{l}{\textit{\textbf{Unified Tokenizer}}} \\
TokenFlow-B$^\dagger$~\cite{geyer2023tokenflow} & 224 & 680 & 32,768 & 1.37 \\
TokenFlow-L$^\dagger$~\cite{geyer2023tokenflow} & 256 & 680 & 32,768 & 0.63 \\
VILA-U$^\dagger$~\cite{wu2024vilau} & 256 & 16$\times$16$\times$4 & 16,384 & 1.80 \\
SemHiTok~\cite{chen2025semhitok} & 256 & 16$\times$16 & 196,608 & 1.16 \\
UniTok$^\dagger$~\cite{ma2025unitok} & 256 & 16$\times$16 & 65,536 & \textbf{0.33} \\
DualToken~\cite{song2025_dualtoken} & 256 & 16$\times$16$\times$4 & - & 0.52 \\ \rowcolor{blue!8}
\modelname (Ours) & 256 & 16$\times$16$\times$4 & 131,072 & \underline{0.43} \\ \bottomrule
\end{tabular}
\end{table}

\subsection{Unified Image Tokenizer}
\label{sec:result_rec}
We benchmark \modelname on the ImageNet-1K validation set to evaluate its reconstruction capability. Table~\ref{tab:main_rfid} reports the commonly used metric rFID. 
Despite training on only 13M images, \modelname achieves a rFID of \textbf{0.43}, substantially outperforming previous methods trained at comparable scale. 
While UniTok~\cite{ma2025unitok} reaches 0.33, it relies on the DataComp-1B~\cite{gadre2023datacomp} corpus with 1.28B image-text pairs. Qualitative reconstruction results on ImageNet-1K are shown in Fig.~\ref{fig:reconstruction_figs}.

\subsection{Unified Understanding and Generation}
\label{sec:result_mllm}
\textbf{Understanding Performance.} Table~\ref{tab:main_understanding} reports the performance achieved by the proposed \modelname and other leading VLMs on diverse visual understanding benchmarks. \modelname achieves leading  performance among discrete unified methods. Specifically,  our model secures the best results on \textbf{7} of 9 evaluated benchmarks, i.e., SEEDBench, GQA, AI2D, RealWorldQA, MMMU, MMBench, and MME, while ranking second-best on MM-Vet and MME-P. Besides, our model demonstrates superior reasoning and comprehensive perception capability. Notably, in reasoning-heavy benchmarks like AI2D and MMMU, \modelname outperforms other discrete models by a large margin (i.e., 76.2 and 45.9, respectively). Furthermore, on the comprehensive MME test, our model (1895.1) significantly surpasses the second-best model in the same category (TokenFlow-B at 1660.4). These results indicate that  the pixel-to-semantic evolution ensures strong semantic representations, making \modelname a competitive unified tokenizer for visual understanding.

\begin{table}[t!]
\centering
\caption{\textbf{Quantitative results on visual understanding benchmarks.} We collect benchmarks including: SEEDBench \cite{seedbench}, MM-Vet \cite{mmvet}, GQA \cite{gqa}, AI2D \cite{ai2d}, RealWorldQA \cite{realworldqa}, MMMU \cite{yue2024mmmu}, MMBench \cite{liu2025mmbench}, MME \cite{fu2023mme} and MME-P (MME-Perception). Three main types of methods are compared: continuous understanding methods (\textit{top}), continuous unified methods (\textit{middle}) and discrete unified methods (\textit{bottom}). The best and the second best results among discrete unified methods are highlighted in \textbf{bold} and \underline{underline} respectively.}
\label{tab:main_understanding}
\resizebox{0.98\linewidth}{!}{%
\begin{tabular}{llcccccccccc}
\toprule
\textbf{Model} & \textbf{LLM} & \textbf{Res.} & \textbf{SEEDB} & \textbf{MM-Vet} & \textbf{GQA} & \textbf{AI2D} & \textbf{RWQA} & \textbf{MMMU} & \textbf{MMB} & \textbf{MME} & \textbf{MME-P} \\ \hline\rowcolor[HTML]{EFEFEF}
\multicolumn{12}{l}{\textit{\textbf{Understanding Only (Continuous)}}} \\
Qwen-VL-Chat~\cite{bai2023qwenvl} & Qwen-7B & 448 & 57.7 & - & 57.5 & - & - & - & - & 1848.3 & 1478.5 \\
VILA~\cite{lin2024vila} & LLaMA-2-7B & 336 & 61.1 & 34.9 & 62.3 & - & - & - & 68.9 & 1533.0 & - \\
ShareGPT4V~\cite{chen2024sharegpt4v} & Vicuna-7B & 336 & 69.7 & 37.6 & 63.3 & 58.0 & 54.9 & 37.2 & 68.8 & 1943.8 & 1567.4 \\
LLaVA-v1.5~\cite{liu2024llava15} & Vicuna-1.5-13B & 336 & 68.1 & 36.1 & 63.3 & 61.1 & 55.3 & 36.4 & 67.7 & 1826.7 & 1531.3 \\ \hline\rowcolor[HTML]{EFEFEF}
\multicolumn{12}{l}{\textit{\textbf{Unified Understanding and Generation (Continuous)}}} \\
Janus~\cite{wu2025janus} & DeepSeek-1.3B & 384 & - & 34.3 & 59.1 & - & - & - & - & 1338.0 &  \\
Unified-IO 2~\cite{lu2024unifiedio2} & 6.8B from scratch & 384 & 61.8 & - & - & - & - & - & 71.5 & - & - \\
LaVIT~\cite{jin2023lavit} & LLaMA-7B & 224 & - & - & 46.8 & - & - & - & - & - & - \\
ILLUME~\cite{wang2025illume} & Vicuna-7B & - & - & 37.0 & - & 71.4 & - & 38.2 & 75.1 & - & 1445.3 \\ \hline\rowcolor[HTML]{EFEFEF}
\multicolumn{12}{l}{\textit{\textbf{Unified Understanding and Generation   (Discrete)}}} \\
QLIP~\cite{zhao2025qlip} & Vicuna-1.5-7B & 392 & - & 33.3 & \textbf{61.8} & - & - & - & - & 1498.3 & - \\
VILA-U~\cite{wu2024vilau} & LLaMA-7B & 384 & 59.0 & 33.5 & 60.8 & - & - & - & - & 1401.8 & - \\
TokLIP~\cite{lin2025toklip} & Qwen2.5-7B-Inst. & 384 & 70.4 & 29.8 & 59.5 & - & - & 43.1 & {\ul 67.6} & - & \textbf{1488.4} \\
TokenFlow-B~\cite{geyer2023tokenflow} & Vicuna-13B & 224 & 60.4 & 22.4 & 59.3 & 54.2 & 49.4 & 34.2 & 55.3 & {\ul 1660.4} & 1353.6 \\
TokenFlow-L~\cite{geyer2023tokenflow} & Vicuna-13B & 256 & 62.6 & 27.7 & 60.3 & 56.6 & 49.2 & 34.4 & 60.3 & 1622.9 & 1365.4 \\
EMU3~\cite{wang2024emu3} & 8B from scratch & - & 68.2 & 37.2 & 60.3 & {\ul 70.0} & {\ul 57.4} & 31.6 & 58.5 & - & - \\
VILA-U~\cite{wu2024vilau} & LLaMA-7B & 256 & 56.3 & 27.7 & 58.3 & - & - & - & - & 1336.2 & - \\
SemHiTok~\cite{chen2025semhitok} & Qwen2.5-7B & 256 & 62.9 & - & 60.3 & - & - & - & - & - & 1355.8 \\
UniTok~\cite{ma2025unitok} & LLaMA-2-7B & 256 & - & 33.9 & {\ul 61.1} & - & - & - & - & 1448.0 & - \\
DualToken~\cite{song2025_dualtoken} & LLaMA-2-7B & 256 & \textbf{71.8} & \textbf{40.5} & - & - & - & {\ul 45.8} & \textbf{74.9} & 1502.7 & - \\ \rowcolor{blue!8}
\modelname (Ours) & Qwen-2.5-7B-Inst. & 256 & \textbf{71.8} & {\ul 39.9} & \textbf{61.8} & \textbf{76.2} & \textbf{62.0} & \textbf{45.9} & \textbf{74.9} & \textbf{1895.1} & {\ul 1455.9}\\ \bottomrule
\end{tabular}
}
\end{table}
\begin{figure}[t!]
    \centering
    \includegraphics[width=\linewidth]{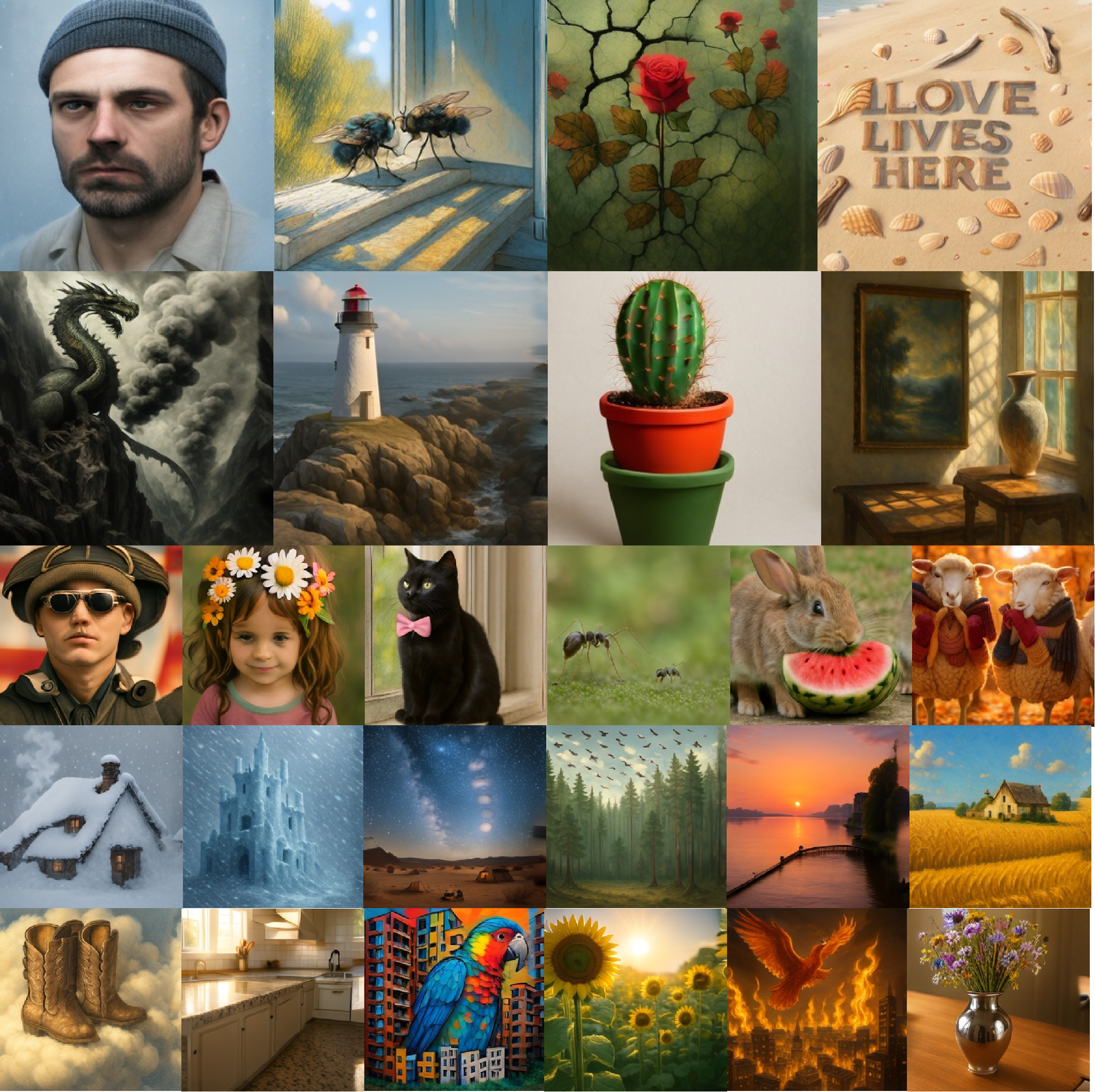}
    \caption{\textbf{Images generated with our unified \modelname at a resolution of 256$\times$ 256.} Our model generates high-quality images spanning diverse visual domains, including portraits, landscapes, objects, artistic paintings, etc.}
    \label{fig:generation_figs}
\end{figure}

\noindent
\textbf{Generation Performance.} 
To evaluate the visual generation capability, we compare our model with other leading methods, including generation-only specialists and unified series, on two standard benchmarks: GenAI-Bench~\cite{jiang2024genai} and GenEval~\cite{geneval}.  As reported in Table~\ref{tab:main_generation}, \modelname demonstrates strong performance, with an overall score of 0.75 on GenEval and 0.87 on GenAI-Bench (Basic). Notably, our model significantly outperforms both unified baselines and specialized diffusion models in complex compositional tasks, particularly in Position (0.69) and Color Attribution (0.62). These results demonstrate that while low-level pixel features successfully evolve into high-level semantics for understanding, the residual trajectory simultaneously preserves granular generative priors, establishing \modelname as a highly balanced and versatile unified tokenizer. Fig.~\ref{fig:generation_figs} shows qualitative image generation results of \modelname. The model demonstrates strong visual generation ability and produces high-quality images across diverse visual domains, including portraits, landscapes, objects, and artistic paintings.

\begin{table}[t!]
\centering
\caption{\textbf{Quantitative results on visual generation benchmarks.} We compare two types of baselines: generation-only specialist models (\textit{top}) and unified MLLMs (\textit{bottom}). \modelname achieves strong performance on both widely used benchmarks, GenAI-Bench~\cite{jiang2024genai} and GenEval~\cite{geneval}.}
\label{tab:main_generation}
\resizebox{0.98\linewidth}{!}{%
\begin{tabular}{llccccccccc}
\toprule
\multirow{2}{*}{\textbf{Model}} & \multirow{2}{*}{\textbf{Type}} & \multicolumn{2}{c}{\textbf{GenAI-Bench~$\uparrow$}} & \multicolumn{7}{c}{\textbf{GenEval~$\uparrow$}} \\ \cmidrule(lr){3-4} \cmidrule(lr){5-11} \cmidrule(lr){9-11}
 &  & Basic & Advanced & \textbf{Overall} & Single Obj. & Two Obj. & Counting & Colors & Position & Color Attri. \\ \hline\rowcolor[HTML]{EFEFEF}
\multicolumn{11}{l}{\textit{\textbf{Geneartion Only}}} \\
SD v2.1~\cite{rombach2022sdv15_21} & Diffusion & 0.78 & 0.62 & 0.50 & 0.98 & 0.51 & 0.44 & 0.85 & 0.07 & 0.17 \\
SD v1.5~\cite{rombach2022sdv15_21} & Diffusion & - & - & 0.43 & 0.97 & 0.38 & 0.35 & 0.76 & 0.04 & 0.06 \\
SDXL~\cite{podell2023sdxl} & Diffusion & 0.83 & 0.63 & 0.55 & 0.98 & 0.74 & 0.39 & 0.85 & 0.15 & 0.23 \\
DALL-E 3~\cite{ramesh2022dalle3} & Diffusion & - & - & 0.67 & 0.96 & 0.87 & 0.47 & 0.83 & 0.43 & 0.45 \\ \hline\rowcolor[HTML]{EFEFEF}
\multicolumn{11}{l}{\textit{\textbf{Unified Understanding and Generation}}} \\
Janus~\cite{wu2025janus} & Autoregressive & - & - & 0.61 & {\ul 0.97} & 0.68 & 0.30 & {\ul 0.84} & 0.46 & 0.42 \\
ILLUME~\cite{wang2025illume} & Autoregressive & 0.75 & 0.60 & 0.61 & \textbf{0.99} & {\ul 0.86} & \textbf{0.45} & 0.71 & 0.39 & 0.28 \\
QLIP-B/16~\cite{zhao2025qlip} & Autoregressive & - & - & 0.48 & 0.91 & 0.59 & 0.22 & 0.80 & 0.17 & 0.24 \\
TokenFlow-XL~\cite{geyer2023tokenflow} & Autoregressive & - & - & 0.63 & 0.93 & 0.72 & \textbf{0.45} & 0.82 & 0.45 & 0.42 \\
EMU3~\cite{wang2024emu3} & Autoregressive & - & - & {\ul 0.66} & \textbf{0.99} & 0.81 & 0.42 & 0.80 & {\ul 0.49} & {\ul 0.45} \\
VILA-U~\cite{wu2024vilau} & Autoregressive & 0.76 & 0.64 & - & - & - & - & - & - & - \\
SemHiTok~\cite{chen2025semhitok} & Autoregressive & 0.83 & 0.65 & - & - & - & - & - & - & - \\
UniTok~\cite{ma2025unitok} & Autoregressive & {\ul 0.85} & \textbf{0.67} & - & - & - & - & - & - & - \\
DualToken-7B~\cite{song2025_dualtoken} & Autoregressive & - & - & \textbf{0.75} & - & - & - & - & - & - \\ \rowcolor{blue!8}
\modelname (Ours) & Autoregressive & \textbf{0.87} & {\ul 0.66} & \textbf{0.75} & \textbf{0.99} & \textbf{0.91} & {\ul 0.44} & \textbf{0.87} & \textbf{0.69} & \textbf{0.62} \\ \bottomrule
\end{tabular}
}
\end{table}

\subsection{Ablation Studies}
\label{sec:resule_ablation}
\noindent
\textbf{Ablation on the Depth of Pixel-to-Semantic Evolution.}
We verify our core design by varying the depth allocation for pixel reconstruction $L_\text{pix}$ and semantic abstraction $L_\text{sem}$. Specifically, we compare three representative configurations: \textbf{1)} an \textit{entangled feature space} where pixel and semantic using  equal depth ($L_\text{pix}=L_\text{sem}=4$), \textbf{2)} a \textit{semantic-to-pixel evolution} where semantic abstraction is performed in earlier stages followed by pixel reconstruction ($L_\text{pix}=16, L_\text{sem}=4$), and \textbf{3)} the proposed \textit{pixel-to-semantic evolution} where low-level pixel features are first modeled and progressively evolved into semantic representations ($L_\text{pix}=4, L_\text{sem}=16$).

As shown in Table~\ref{tab:ab1_depth}, the results reveal two critical insights regarding the unified latent space:
\textbf{First}, the \textbf{\textit{entangled feature space}} (\textit{Row 1}) leads to degraded performance on both reconstruction and semantic benchmarks, indicating that directly mixing pixel and semantic features within the same trajectory causes severe representation interference.
\textbf{Second}, the \textbf{\textit{semantic-to-pixel evolution}} (\textit{Row 2}) achieves the best reconstruction quality but yields noticeably weaker results on both understanding and generation tasks, indicating that high-level semantics cannot be effectively regressed from pixel-heavy tokens. 

In contrast, the proposed \textbf{\textit{pixel-to-semantic evolution}} (\textit{Row 3}), where low-level pixel features progressively evolve into high-level semantic representations, produces the most balanced performance across reconstruction, understanding, and generation. 
These results demonstrate that the pixel-to-semantic evolution effectively reconciles the representational requirements of pixel-level generation and semantic-level understanding. By establishing this cascaded multi-stage progression, \modelname mitigates the typical performance degradation seen in entangled architectures, achieving a unique synergy between generative quality and semantic understanding capability.

\begin{figure}[t!]
    \centering
    
    \begin{subfigure}{0.49\linewidth}
        \centering
        \includegraphics[width=\linewidth]{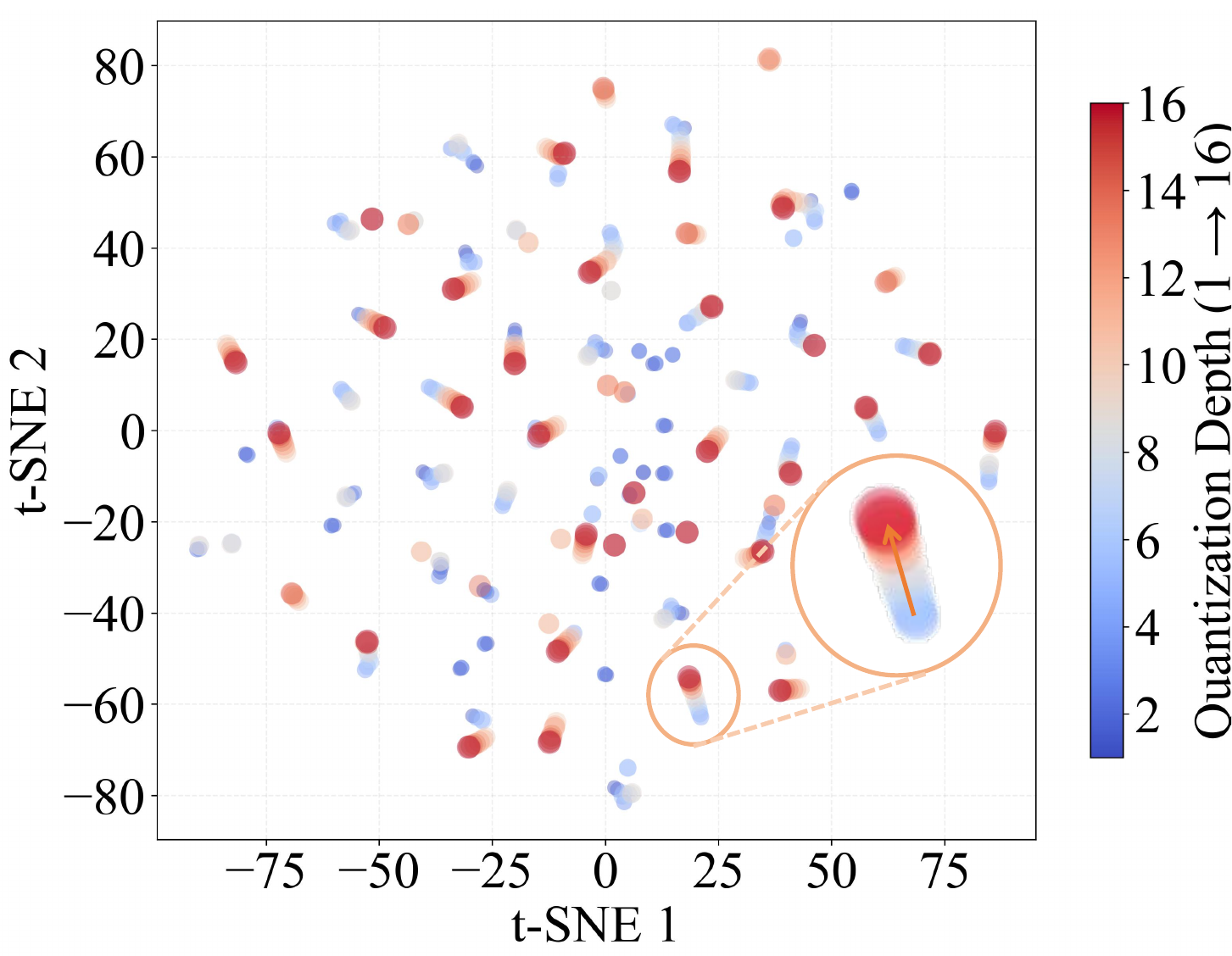}
        \caption{\textbf{Latent Evolution Trajectory.} t-SNE visualization reveals a continuous transition from pixel-level perception (blue) to high-level semantics (red) within a single, unified latent space, bridging the gap between generation and understanding.}
        \label{fig:ab3a_trajectory_tsne}
    \end{subfigure}
    \hfill
    \begin{subfigure}{0.49\linewidth}
        \centering
        \includegraphics[width=\linewidth]{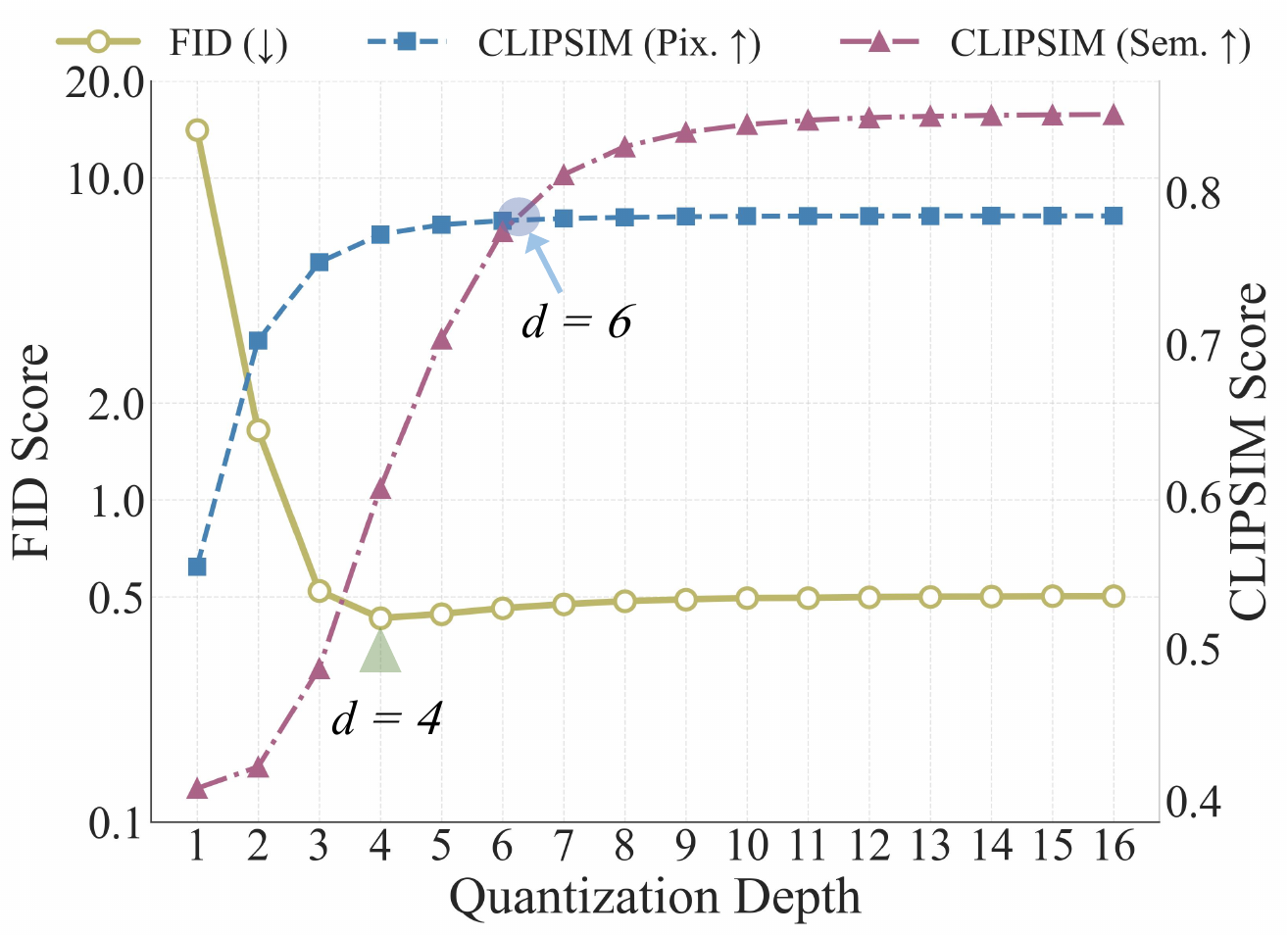}
        \caption{\textbf{Feature Refinement Along the Evolution Trajectory.} At early depths (1–4), representations capture fine-grained pixel structures for reconstruction. As depth increases (8–16), the cumulative residuals progressively encode abstract concepts, achieving seamless alignment with high-level semantic features.}
        \label{fig:ab3b_trajectory_depth}
    \end{subfigure}
    
    \caption{\textbf{Analysis of the Unified Latent Space.}}
    \label{fig:ab3_trajectory_analysis}
\end{figure}

\noindent
\textbf{Analysis of the Latent Evolution Trajectory.}
Fig.~\ref{fig:ab3a_trajectory_tsne} provides qualitative evidence of the shared latent space in \modelname. The t-SNE visualization reveals a continuous progression along the residual stack, transitioning from pixel-level perception (blue) at shallower depths to high-level semantics (red) at deeper stages. The trajectory of each token (as indicated by the magnified inset) demonstrates that instead of maintaining decoupled representations, \modelname effectively unifies generative and discriminative features within a single, evolving latent path. This observation confirms that the residual quantization layers progressively refine low-level visual representations into high-level semantic concepts, providing a unified representation for both understanding and generation.
\begin{table}[t!]
\centering
\caption{\textbf{Ablation on unified feature trajectories.} Three different unified strategies are compared: 1) entangled pixel and semantic modeling, 2) semantic-to-pixel evolution, and 3) our proposed pixel-to-semantic evolution. Our method achieves the most balanced performance across all tasks.}
\label{tab:ab1_depth}
\resizebox{0.98\linewidth}{!}{%
\begin{tabular}{ccccccccccc}
\toprule
\multirow{2}{*}{$L_\text{pix}$} & \multirow{2}{*}{$L_\text{sem}$} & \multicolumn{3}{c}{\textbf{Reconstruction}} & \multicolumn{3}{c}{\textbf{Understanding}} &  & \textbf{Generation} &  \\ \cmidrule(lr){3-5} \cmidrule(lr){6-8} \cmidrule(lr){9-11}
 &  & rFID~$\downarrow$ & PSNR~$\uparrow$ & SSIM~$\uparrow$ & SEEDB~$\uparrow$ & GQA~$\uparrow$ & MME~$\uparrow$ & GenAI (\textit{Bas.})~$\uparrow$ & GenAI (\textit{Adv.})~$\uparrow$ & GenEval~$\uparrow$ \\ \hline
4 & 4 & 0.66 & 22.96 & 0.76 & 62.7 & 58.0 & 1668.9 & 0.70 & {\ul 0.59} & {\ul 0.64} \\
16 & 4 & \textbf{0.44} & \textbf{24.39} & \textbf{0.79} & {\ul 64.6} & {\ul 60.9} & {\ul 1731.6} & {\ul 0.71} & 0.58 & 0.60 \\ \rowcolor{blue!8}
4 & 16 & {\ul 0.55} & {\ul 23.60} & {\ul 0.77} & \textbf{67.1} & \textbf{61.3} & \textbf{1793.5} & \textbf{0.75} & \textbf{0.62} & \textbf{0.67} \\ \bottomrule
\end{tabular}
}
\end{table}
\begin{figure}[t!]
    \centering
    \includegraphics[width=\linewidth]{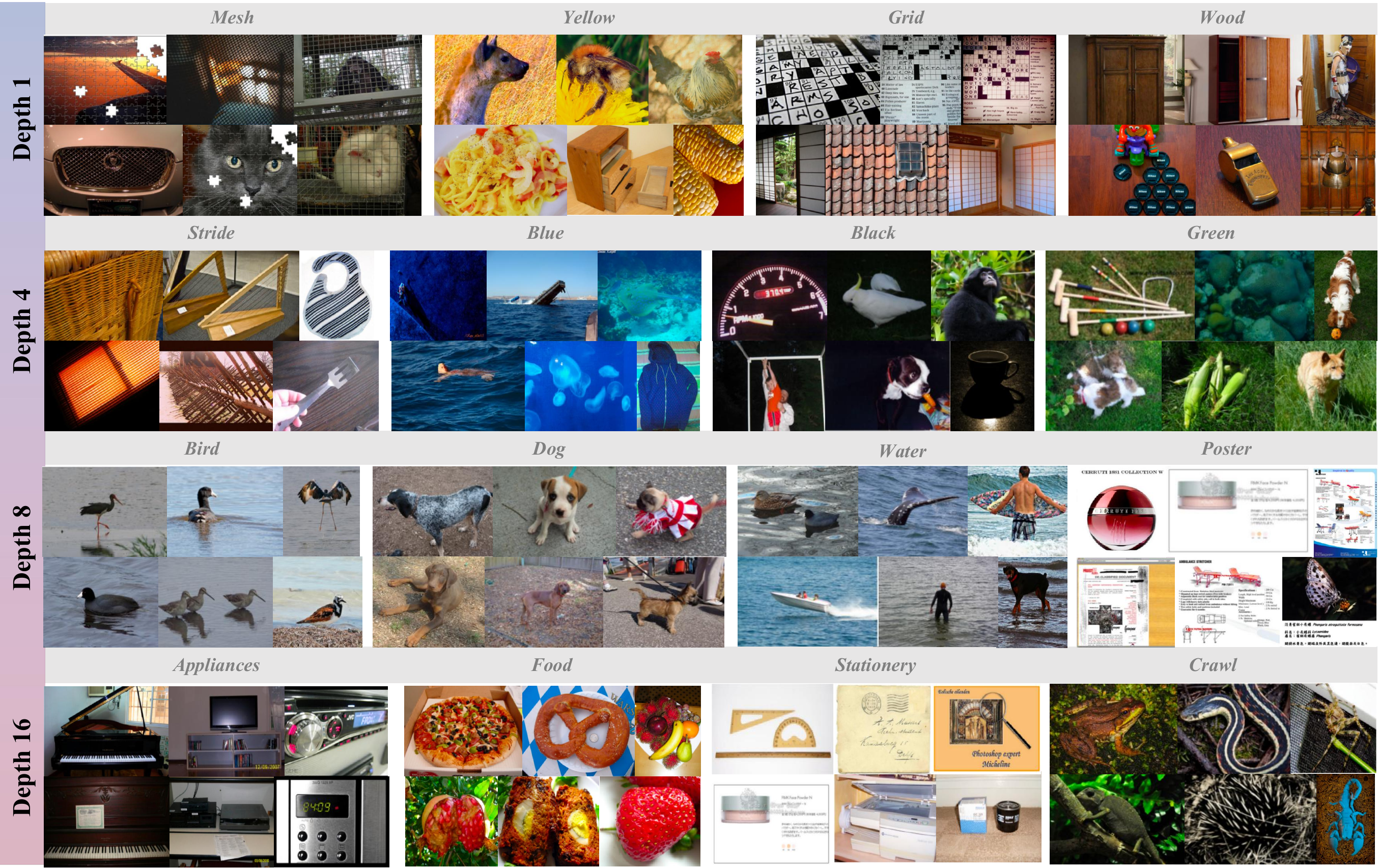}
    \caption{\textbf{Visualization of the feature evolution trajectory across different residual depths.} We visualize image clusters grouped by K-means centroids at specific depth stages $L \in \{1, 4, 8, 16\}$. (Top Rows) At shallow depths ($L=1, 4$), the latent features primarily capture \textit{perceptual primitives} such as textures (\textit{Mesh}, \textit{Grid}), color palettes (\textit{Yellow}, \textit{Blue}), and low-level structures (\textit{Stride}). (Bottom Rows) As depth increases ($L=8, 16$), the representation \textit{naturally evolves} toward \textit{high-level semantic abstractions}, grouping diverse instances into coherent conceptual categories (e.g., \textit{Appliances}, \textit{Food}, \textit{Crawl}).
    }
    \label{fig:suppl_vis_clusters}
\end{figure}

\noindent
\textbf{Analysis of Feature Refinement Along the Evolution Trajectory.}
We quantitatively analyze the representation evolution across the residual stack in Fig.~\ref{fig:ab3b_trajectory_depth}. Specifically, we monitor three key metrics along the trajectory: \textbf{1)} rFID of the reconstructed image; \textbf{2)} $\text{CLIPSIM}_{\text{pix}}$, calculated as the semantic similarity between the reconstructed image and the original input; and \textbf{3)} $\text{CLIPSIM}_{\text{sem}}$, which measures the semantic similarity between decoded semantic features and the GT semantic features. These metrics allow us to observe how visual information evolves between explicit pixel reconstruction and semantic latent encoding.

The empirical trends reveal a clear functional divergence. At early depths ($d \le 4$), $\text{CLIPSIM}_{\text{pix}}$ and rFID improves rapidly, indicating that the initial residuals are highly efficient at capturing the primary semantic structures visible at the pixel level. However, beyond $d=4$, $\text{CLIPSIM}_{\text{pix}}$ enters a plateau, suggesting a representational bottleneck where further pixel-level refinement no longer yields additional semantic gain. In contrast, $\text{CLIPSIM}_{\text{sem}}$ continues to ascend significantly. This decoupling of trends demonstrates that the later residual stages successfully transcend the limitations of raw pixel alignment, shifting their capacity toward encoding higher-order semantic concepts.

We further \textbf{\textit{qualitatively}} investigate the representational characteristics of features at different quantization depths $L \in \{1, 4, 8, 16\}$. Specifically, we extract the latent features from each specified depth and perform K-means clustering on a subset of the ImageNet-1K validation set. Fig.~\ref{fig:suppl_vis_clusters} visualizes the images closest to the cluster centroids, revealing a clear transition in semantic granularity along the residual trajectory.
At the initial stages ($L=1, 4$), the clusters are primarily dominated by \textit{perceptual primitives} and low-level structural cues. As shown in the top rows of Fig.~\ref{fig:suppl_vis_clusters}, the model groups images based on shared textures (e.g., \textit{Mesh}, \textit{Grid}), dominant color palettes (e.g., \textit{Yellow}, \textit{Blue}), or local geometric patterns (e.g., \textit{Stride}). This observation confirms that earlier residual layers focus on capturing fine-grained pixel-level fidelity, which is essential for preserving structural integrity during image reconstruction and generation tasks.
As the residual depth increases to $L=8$ and $L=16$, a significant shift toward \textit{high-level semantic abstraction} is observed. At $L=8$, the clusters begin to reflect broad taxonomic categories such as \textit{Bird}, \textit{Dog}, and \textit{Water}, indicating a transition from local textures to object-level awareness. By $L=16$, the features exhibit robust semantic consistency, successfully grouping complex concepts like \textit{Appliances}, \textit{Stationery}, and \textit{Crawl} (e.g., reptiles and insects) despite vast intra-class variations in scale, pose, and background. 

This evolutionary trajectory provides strong evidence that our tokenizer naturally achieves task decoupling within a unified latent space: the shallow layers prioritize the perceptual details required for high-fidelity generation, while the deeper layers refine the representation into the semantic abstractions necessary for visual understanding.

\noindent
\textbf{Effectiveness of Feature Decoupling Strategy.} 
Table~\ref{tab:ab2_decouple_arch} compares the impact of four decoupling strategies: \textbf{1)} \textit{entangled features} (Fig.~\ref{fig:teaser}(a), e.g., ViLA-U~\cite{wu2024vilau}) directly shares the same pixel and semantic features, implemented as the configuration in Table~\ref{tab:ab1_depth} with $L_\text{pix}=L_\text{sem}=4$; \textbf{2)} \textit{Layer-wise decoupled features from a single encoder} (Fig.~\ref{fig:teaser}(b), e.g., DualToken~\cite{song2025_dualtoken}) separates semantic and pixel representations by extracting them from different depths of a shared vision encoder, whereas our method derives both representations from the top-layer features without splitting intermediate layers; \textbf{3)} \textit{Encoder-level decoupled features} (Fig.~\ref{fig:teaser}(b), e.g., TokenFlow~\cite{geyer2023tokenflow}) decouples pixel and semantic representations using two independent visual encoders, whereas our method uses a shared encoder and evolves both features within a unified latent space.
and \textbf{4)} our pixel-to-semantic evolution features (Fig.~\ref{fig:teaser}(c)).

\begin{wraptable}{r}{5cm}
\vspace{-3em}
\centering
\caption{\textbf{Ablation on different feature decoupling method.}}
\label{tab:ab2_decouple_arch}
\begin{tabular}{lc}
\toprule
\textbf{Decouple Method} & \textbf{rFID~$\downarrow$} \\ \hline
1) No decouple & 0.66 \\
2) Different feature layer & 0.59 \\
3) Different branch & 0.68 \\ \rowcolor{blue!8}
4) Ours & \textbf{0.55} \\ \bottomrule
\end{tabular}
\end{wraptable}

As shown in Table~\ref{tab:ab2_decouple_arch}, entangling pixel and semantic features (0.66) or using separate encoder branches (0.68) leads to inferior reconstruction quality, while decoupling features from different layers of a single encoder improves performance (0.59), our pixel-to-semantic evolution achieves the best rFID of 0.55. This demonstrates that our evolutionary approach preserves fine-grained visual details more effectively, compared to rigid decoupling or fully entangled representations.

\section{Conclusion}
This paper presents \modelname, a unified image tokenizer designed to support both visual understanding and generation. 
Existing approaches typically either entangle semantic and pixel representations in a shared feature space or overly decouple them into independent branches, which limits the effectiveness of unified multimodal modeling. 
To address this issue, \modelname represents images as a residual evolution trajectory within a shared latent space, where pixel-level representations progressively evolve into higher-level semantic abstractions through cascaded residual quantization. 
This design enables task-decoupled representations for generation and understanding while preserving their intrinsic consistency. Extensive experiments demonstrate that \modelname achieves strong performance on both visual generation and understanding benchmarks, despite being trained on significantly less data than prior unified models. 
We hope the proposed residual evolution paradigm can offer new insights into tokenizer design and facilitate the development of more capable unified multimodal systems.


\bibliographystyle{plainnat}
\bibliography{references}


\end{document}